\documentclass{article}

\usepackage{PRIMEarxiv}

\usepackage[utf8]{inputenc} 
\usepackage[T1]{fontenc}    
\usepackage{hyperref}       
\usepackage{url}            
\usepackage{booktabs}       
\usepackage{amsfonts}       
\usepackage{nicefrac}       
\usepackage{microtype}      
\usepackage{lipsum}
\usepackage{fancyhdr}       
\usepackage{graphicx}       
\graphicspath{{media/}}     

\usepackage{lineno}
\usepackage{amsmath}
\usepackage{amssymb}
\usepackage{algorithm}
\usepackage{algpseudocode}
\usepackage{subcaption}
\usepackage{tikz}
\usepackage{booktabs}
\usepackage{multirow}
\usepackage{tabularx}

\modulolinenumbers[5]

\title{Closing the Gap Between SGP4 and High-Precision Propagation via Differentiable Programming
}

\author{
  Giacomo Acciarini \\
  University of Surrey \\
  \texttt{g.acciarini@surrey.ac.uk} \\
   \And
  Atılım Güne\c{s} Baydin \\
  University of Oxford \\
  \texttt{gunes@robots.ox.ac.uk} \\
  \And
  Dario Izzo \\
  European Space Agency \\
  \texttt{dario.izzo@esa.int}
}

\begin{document}
\let\oldhat\hat
\renewcommand{\vec}[1]{\pmb{\mathrm{#1}}}
\renewcommand{\hat}[1]{\oldhat{\pmb{\mathrm{#1}}}}
\maketitle

\begin{abstract}
The simplified general perturbations 4 (SGP4) orbital propagation is one of the most widely used approaches for computing a rapid and relatively reliable prediction of the positions and velocity of objects orbiting Earth.
Over time, SGP models have undergone refinement to enhance their efficiency and accuracy. Yet, they still fall short of the precision offered by high-precision numerical propagators, which can predict the positions and velocities of space objects in low-Earth orbit with significantly smaller errors.
In this study, we introduce a novel differentiable version of SGP4, named $\partial$SGP4. By reconfiguring SGP4 into a differentiable program using PyTorch, we unlock its differentiability for various space-related applications, encompassing spacecraft orbit determination, state conversion, covariance similarity transformation, state transition matrix computation, covariance propagation, and more.
A side advantage of this PyTorch-based model is its capacity for "embarrassingly parallel" orbital propagation across batches of TLEs. It can harness the processing power of CPUs, GPUs, and advanced hardware alike, to be able to predict batches of satellites at future times in a distributed fashion. Furthermore, the inherent differentiability of $\partial$SGP4 renders it compatible with modern machine learning methodologies.
Consequently, we propose a novel orbital propagation paradigm, ML-$\partial$SGP4. In this paradigm, the orbital propagator is imbued with neural networks. Through stochastic gradient descent during the learning process, the inputs, outputs, and parameters of this combined model can be iteratively refined to achieve precision surpassing that of SGP4. Fundamentally, the neural networks function as identity operators when the propagator adheres to its default behavior as defined by SGP4. However, owing to the differentiability ingrained within $\partial$SGP4, the model can be fine-tuned with ephemeris data to learn corrections to both inputs and outputs of SGP4. This augmentation enhances precision while maintaining the same computational speed of $\partial$SGP4 at inference time.
This empowers satellite operators and researchers, equipping them with the ability to train the model using their specific ephemeris or high-precision numerical propagation data.
\end{abstract}

\keywords{SGP4 \and Orbital Propagation \and Differentiable Programming \and Machine Learning \and Spacecraft Collision Avoidance \and Kessler \and Kessler Syndrome \and AI for Space \and Applied Machine Learning for Space
}

\section{Introduction}
\label{sec:introduction}

As the number of resident space objects (RSO) increases, space situational awareness (SSA) becomes a critical aspect of today's satellite operations, involving the characterization, detection, and tracking of objects in orbit around Earth. The conspicuous space debris population poses a threat to the increasing number of operational satellites: collisions among space objects have catastrophic consequences for the safety of surrounding objects and for the future of space and the accessibility to certain orbital regimes. An essential building block to ensure safe operations in space is orbital propagation, which plays a key role in SSA since it is used to predict and catalog the movement of RSOs in space. These propagators take an initial state, describing the position and velocity of a space object, and integrate them at future times using a model of the forces that act on the object in the near-Earth environment. Some examples of these forces can include the gravitational force exerted by the Earth and other Solar System bodies, as well as the atmospheric drag and the solar radiation pressure. Depending on the perturbations that are taken into account and on the numerical scheme used to tackle the integration, the accuracy and speed of these methods in solving the set of ordinary differential equations describing the satellites' motion can substantially vary. 

Typically, we distinguish four types of orbital propagators~\cite{vallado2001fundamentals}. \textit{General perturbation techniques}: these produce analytical solutions to the equations of motion. These solutions are very fast to evaluate but are only an approximation that is valid under some assumptions~\cite{brouwer_1959, lyddane1963small}. \textit{Special perturbation techniques}: these consist of the numerical integration of the equations of motion with the necessary perturbing accelerations. Typically, these methods rely on some numerical scheme for the integration (e.g. Taylor method, Runge-Kutta methods, Adams–Bashforth-Moulton methods, etc.), which leads to slower but more accurate solutions w.r.t. general perturbation techniques~\cite{montenbruck2002satellite}. \textit{Semianalytical techniques}: these try to combine the accuracy of numerical methods with the speed of analytical techniques. Often, they consist of numerical schemes that solve a set of equations that are derived from the original equations of motion under certain assumptions, which enable faster integration~\cite{liu1980semianalytic}. \textit{Hybrid techniques}: these are an emergent trend that tries to combine the speed and accuracy of the above method by using statistical and machine learning methods. In this paradigm, the prediction of general perturbation methods is corrected via a machine learning and/or statistical model, to more closely resemble the accuracy of the numerical propagator prediction~\cite{perez2013application, san2018hybrid}.

During the 1960s, the US Air Force initiated an endeavor to develop orbital propagation models that could provide reliable and rapid predictions of the positions of resident space objects based on initial position and velocity data. This initiative led to the creation of what are known as simplified perturbation models. One of the most popular general perturbation techniques is the Simplified General Perturbations 4 (SGP4) method, which belongs to the class of the simplified perturbations models (SGP, SGP4, SDP4, SGP8, SDP8)~\cite{kozai1959motion}. This consists of a mathematical model with a long history of development (starting from the 1960s) used to predict the position and velocity of an Earth-orbiting object~\cite{lane1965development, vallado2006revisiting}, although the first release of the refined SGP4 propagator only happened in the 1970s~\cite{hoots1980spacetrack}. Originally, it was only applicable to satellites whose orbital period is less than 225 minutes, although an extension of this model for higher periods has also been developed, which is known as SDP4. It owes its popularity to the fact that simplified perturbation models are the only ones that support two-line elements (TLEs) data in their original format: these are the only openly available ubiquitous source of information for the position and velocity of satellites. TLE is a data format that describes in two lines of 69 characters each, the position, velocity, characteristics (e.g. drag term), and other information (e.g. satellite ID, etc.) of RSOs around the Earth. United States Space Force routinely tracks objects in space and publicly releases their TLE elements through the Space-Track website~\footnote{\url{https://www.space-track.org/auth/login}, date of access: March 2023.}\footnote{\url{https://celestrak.org/NORAD/elements/}, date of access: March 2023.}. Whilst TLE data are released publicly, they are created from a mixture of analytical theories, observations and coordinate systems that are not revealed~\cite{vallado2006revisiting}. For example, SGP4 uses power density functions to estimate the drag force that acts on the satellite motion~\cite{cranford1969improved, lane1979general}: these require the knowledge of a term that is related to the ballistic coefficient, called Bstar parameter. Since a mix of simplified perturbations theory and observational data (processed through batch least squares) is used to produce a single TLE, the Bstar parameter often absorbs force model errors~\cite{vallado2001fundamentals}.

All these unrevealed underlying processes necessary to produce a TLE make the transformation from TLE to osculating elements not possible without errors/assumptions and their use is therefore recommended only with a simplified perturbations propagator. Typical accuracies of these propagators with TLE data are in the order of kilometers for predictions within a week, compared to the ground truth states of the object~\cite{kelso2007validation, flohrer2008assessment, dong2010accuracy, vallado2012two}. For this reason, when osculating orbital elements data are available, special perturbation techniques are often preferred in applications that require high accuracies (e.g. collision avoidance). This has motivated several researchers to improve the SGP4 prediction via statistical and machine learning methods, to make it competitive with special perturbation techniques while retaining its computational advantages. In general, there is a plethora of studies on orbit prediction enhancements using Gaussian processes, state vector machines, and other machine learning techniques~\cite{peng2020machine}. In the context of SGP4, most of the proposed techniques focus on creating datasets of errors between SGP4 predicted states and ephemeris data (or high accuracy numerical propagators), which are then used to train machine learning models that can learn to correct these errors~\cite{peng2023medium}. Recently, some researchers have also devised a fusion strategy between orbit prediction and Bayesian inference methods to enhance the SGP4 predictions from TLE data~\cite{peng2021fusion}. 

Motivated by these works, and by the lack of a general framework that allows improving SGP4 predictions integrating highly accurate data and/or predictions, we developed a differentiable SGP4 ($\partial$SGP4) version on Python, which we make available through an open-source repository on the European Space Agency Github organization\footnote{\url{https://github.com/esa/dSGP4}, date of access: March 2023.}. Our contributions consist of first re-writing the SGP4 propagator to make it compatible with modern programming techniques, which enable automatic differentiation and "embarrassingly parallel" computations via CPU and/or GPU. We then show that having an SGP4 propagator that is differentiable with respect to its inputs opens the possibility of using the Jacobian matrix for several applications (e.g. state covariance propagation, state transition matrix computation, gradient-based optimization). This has also consequences in many fields related to SSA, such as orbit determination, among others. To underscore the significance of employing $\partial$SGP4 in practical contexts, we reference a work by Vallado on SGP4~\cite{vallado2008sgp4}, where he suggests that while elegant operational systems typically opt for numerical integration, a straightforward analytical propagation theory like SGP4 might justify employing finite differencing. According to him, the effort and resources invested in developing partial derivatives for various data types and state forms to achieve computational efficiency are often outweighed by the simplicity of incorporating an additional iteration or two with a finite differencing approach. Through the use of $\partial$SGP4, we demonstrate that the computation of partial derivatives can be seamlessly executed without incurring any extra time or financial costs, thanks to automatic differentiation. Additionally, this approach mitigates local truncation errors associated with finite differencing.

Finally, we also discuss the applications of such a differentiable SGP4 model in a new hybrid propagator paradigm, where the machine learning model is made of an initial neural network that ingests the data, which is then passed through $\partial$SGP4, and finally through another neural network. In this way, owing to the differentiability of $\partial$SGP4, it is possible to take gradients of the entire model, therefore applying standard stochastic gradient descent techniques and training $\partial$SGP4 together with neural network models. This framework enables backpropagating through the whole model, and learning corrections of SGP4 inputs and outputs to match the predicted state: in practice, this means that the neural networks, if successful, learn the differential correction terms needed to correct the low-accuracy predictions of SGP4, to better match either observations or high-fidelity propagations, and retain the advantageous SGP4 speed.  Thus, for the first time, both inputs and outputs (and potentially hyperparameters of SGP4) can be learned for the given prediction task, to maximize the accuracy of the predicted orbital states, while maintaining a fast and parallelizable propagator. 

The paper is organized as follows: in Section~\ref{sec:related_work}, we discuss relevant background on the SGP4 model and differentiable programming. Then, in Section~\ref{sec:dSGP4_an_open_source_differentiable_sgp4_program}, we discuss the features of the propagator, including automatic differentiation and batch mode. In Sec.~\ref{sec:mldSGP4}, we discuss its machine learning integration, by both outlining the general framework and discussing simulation results that include the use of the machine learning $\partial$SGP4 framework to improve the SGP4 predictions on a batch of SpaceX Starlink satellites. While in Sec.~\ref{sec:satellite_orbit_determination}, we introduce the use of $\partial$SGP4 in the context of satellite orbit determination. Other applications including gradient-based optimization, osculating to TLE elements conversion,  covariance propagation, similarity transformation, state transition matrix computation, and batch propagation are included as tutorials of the open-source library and released with $\partial$SGP4. Then, we also conduct an experiment on a group of SpaceX Starlink satellites where $\partial$SGP4 and ephemeris data are used to train the ML-enhanced hybrid propagator, to improve the baseline SGP4 predictions. Finally, in Section~\ref{sec:conclusions}, we highlight the conclusions of our work.
\section{Related Work}
\label{sec:related_work}
\subsection{SGP4 Model}
The Simplified General Perturbations-4 (SGP4)~\cite{vallado2006revisiting} software is an orbit propagation model and it is the only one recommended to be used in combination with two line elements (TLEs) orbit data. Two-line elements are a very popular data format used for space object cataloging. They encapsulate information associated with the satellite position and velocity (using general perturbation mean orbital elements) at a given epoch, together with some identifiers for the satellite (e.g. satellite number, international designator, etc.)~\cite{vallado2001fundamentals}. Six quantities in this format (the mean motion $n$, the eccentricity $e$, the inclination $i$, the right ascension of the ascending node $\Omega$, and the mean anomaly $M$) are associated with the position and velocity in space of the object, while the other three (i.e., the mean motion rate $\dot{n}$, mean motion acceleration $\ddot{n}$, drag-like parameter $B^*$) are used to describe the effect of perturbations in the satellite motion, and the time in UTC is used to indicate the epoch at which the TLE is generated. All these two-line elements variables are directly used by the SGP4 model, which then produces ephemerides for the resident space object at the required time, in the true equator, mean equinox (TEME) coordinate system~\cite{vallado2001fundamentals}. 

\subsection{Differentiable Programming}
Differentiable programming is a programming paradigm in which models can be built through a computer program and then their parameters can be differentiated through automatic differentiation (AD). This allows users to perform gradient-based optimization of the model outputs with respect to its parameters. This means that users can leverage gradient-based techniques to optimize large and complex models with many parameters, adjusting those parameters in order to minimize the deviation of the model output with respect to ground truth data. With the advent of artificial intelligence and neural network models, this paradigm has become increasingly popular, and many scientific fields have successfully used differentiable programming in several areas: from computer science to robotics, ray tracing, image processing, probabilistic programming, electronics, etc.~\cite{degrave2019differentiable, li2018differentiable, hu2019difftaichi,izzo2017differentiable}

\section{\texorpdfstring{$\partial$}{partial}SGP4: an Open-Source Differentiable SGP4 Program}
\label{sec:dSGP4_an_open_source_differentiable_sgp4_program}
\subsection{\texorpdfstring{$\partial$}{partial}SGP4 Overview: Automatic Differentiation and Batch Mode}
\label{sec:automatic_differentiation}
Automatic differentiation is a technique used for computing and evaluating derivatives of functions. It can be applied to any function that can be represented as a sequence of elementary operations, whose derivatives are known~\cite{griewank2008evaluating}. The basic idea of AD is to build a graph of elementary operations with known derivatives, which is then used (either in forward or reverse mode) to precisely compute the derivative at a given point. Before automatic differentiation, derivatives could either be computed manually, with a time-consuming and error-prone procedure, through symbolic differentiation, or through numerical differentiation. This last alternative is not ideal due to the presence of truncation and round-off errors, which might substantially jeopardize the accuracy of the found derivatives, if not chosen carefully. On the other hand, symbolic differentiation would allow to access the derivative of the function for the entire domain, but it can result in the size of the expressions growing exponentially (something known as "expression swell"), as the number of variables and operations is increased, due to the product rule of derivative~\cite{corliss1988applications}. Automatic differentiation solves these issues, by returning the exact value (up to machine precision) of the derivatives of a computer program evaluated at a certain point, without incurring exponentially growing expressions~\cite{baydin2018automatic}. The drawback, besides the fact that it is not applicable for non-differentiable functions, is that this is only the derivative evaluated at a given point, and one does not have access to the symbolic form of the derivative across the entire domain. The differentiability constraint implies that the computer program needs to comply with certain requirements and does not admit discontinuities, singularities, or non-differentiable features. In this case, we have implemented $\partial$SGP4 using the Python library Pytorch~\cite{paszke2019pytorch}, which is one of the most popular open-source ML libraries that supports AD in use in deep learning research and applications. To ensure the fidelity of $\partial$SGP4, our approach encompassed several key verification steps.
First of all, we validated the $\partial$SGP4 model's outputs against the "Revisiting Spacetrack Report \#3" SGP4 model~\cite{vallado2006revisiting}. This involved the design and execution of continuous integration tests (that can be found in the open-source library) to confirm that all mirrored functionalities faithfully reproduced the behavior of the original implementation\footnote{\url{https://github.com/esa/dSGP4/tree/master/tests}, date of access: February 2024.}. As a part of these tests, we also verified that we could reproduce the output of the Space Track official SGP4 release\footnote{\url{https://www.space-track.org/documentation\#/sgp4}, date of access: February 2024.}: this was done by propagating several different TLEs at different future times (between a few hours to a few days), and checking that the results of the propagation match up to below 1 mm in position and 1 mm/s in velocity the official release results. Our verifications also extended to the examination of automatic differentiation outputs for the input variables. To ensure correctness, we compared (up to local truncation errors caused by the finite difference technique) that the results obtained from $\partial$SGP4 automatic differentiation functionality match the ones found via finite differencing of the original SGP4.

\begin{figure}[htb!]
     \centering
    \includegraphics[width=\textwidth]{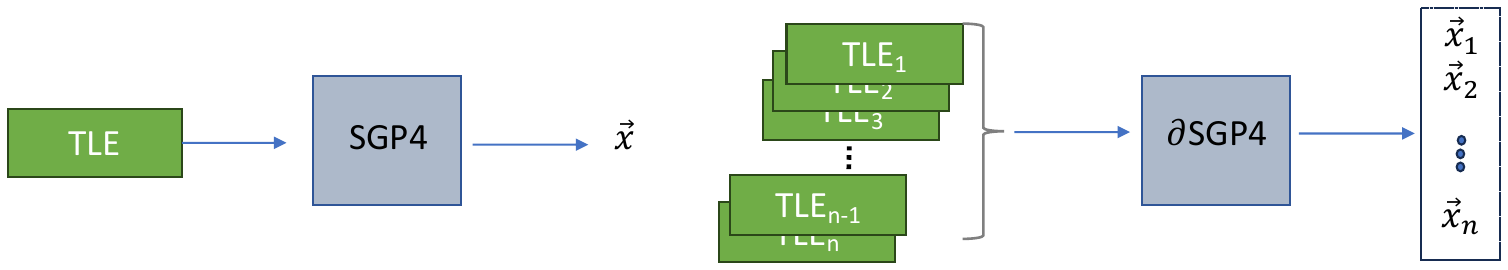}
\caption{Schematic illustration of batch mode supported by $\partial$SGP4 (on the right) vs the standard SGP4 implementation (on the left).}
\label{fig:batch_mode}
\end{figure}

As previously discussed, $\partial$SGP4 also offers an additional capability, as we show in Figure~\ref{fig:batch_mode}. While retaining the traditional functionality of predicting object states from a TLE for specific times or multiple times, $\partial$SGP4 offers a distinctive feature called batch mode. In this mode, batches of TLEs can be provided to the model, which in turn returns the corresponding batch of Cartesian states. This expanded functionality has also undergone comprehensive testing within our library.

Differentiable physics simulators have the key advantage that can be run both in forward and backward mode. When the forward mode is active, the simulator has the commonly known behavior of returning a certain output from the provided inputs: however, when the backward mode is activated, the model can also return the gradients of the outputs with respect to any input and parameter of the model. This allows the integration of this simulator with gradient-based methods, where the gradient information can be leveraged to learn and control the physical system. This approach enables a variety of inverse problems that involve orbital propagation, which we will also discuss in the coming sections (e.g. satellite orbit determination). The interested reader can also find in the $\partial$SGP4 website~\footnote{\url{https://esa.github.io/$\partial$SGP4}, date of access: October 2023.} a plethora of tutorials on how to apply the $\partial$SGP4 framework to common gradient-based tasks for SSA applications (e.g. state transition matrix and sensitivity matrix computation, covariance propagation, and gradient-based optimization). 
The simulator is treated as a differentiable function, and a certain objective function is minimized, which depends on the simulator. Mathematically this is formulated as a minimization problem of a certain objective function: $J(\vec{\theta})$, where $\vec{\theta}$ is an $n$-dimensional vector of model parameters. The goal is to update the parameters in such a way that $J(\vec{\theta})$ reaches the minimum value: this is done in the direction opposite of the gradient of the objective function w.r.t. the parameters. In its simplest (i.e., vanilla gradient descent) form this can be written as $\vec{\theta}_{i+1} = \vec{\theta}_i - \eta \cdot \nabla_{\theta} J(\vec{\theta}_i)$, where $\eta$ is known as learning rate~\cite{ruder2016overview}. Therefore, it is clear that by formulating the orbital propagator as a differential program, its integration with any gradient-related task, such as vanilla gradient descent is possible. This also has the direct consequence of being able to use these gradients within root-finding algorithms that leverage gradient information.

Beyond the advantages of parallel computing, gradient computation, and more, the differentiability of SGP4 opens the door to a new propagator paradigm. This paradigm involves integrating a low-fidelity propagator with neural networks to jointly correct inputs and outputs of SGP4, in a supervised learning prediction task enhancing model accuracy while retaining computational efficiency. To our knowledge, this is the first time in which an orbital propagator is backpropagated together with other machine learning architectures, to combine the stochastic gradient descent's capability of reducing the prediction error, with the physics-based prediction of the model.

\subsection{Performance Profiling}

As already discussed in the previous section, an important advantage of having a differentiable SGP4 program written in PyTorch is the possibility to massively distribute the propagation of TLEs, which we refer to as batch-mode propagation, leveraging either multiple CPUs or GPU acceleration (through CUDA). In this way, we enable the possibility to have embarrassingly parallel computations of multiple TLEs at multiple times, through hardware parallelization. To quantitatively assess this capability, we display in Tab.~\ref{tab:dSGP4_speed} the performance of the simple SGP4, against $\partial$SGP4 on 1 CPU and GPU. We compare the results using float64 numbers (this is not ideal for the GPU case, but it might be important for orbital propagation applications). In terms of CPU processor, we have used the MacBook M1 Pro processor, while in terms of GPU, we have used the NVIDIA V100 Tensor Core GPU\footnote{\url{https://www.nvidia.com/en-us/data-center/v100/}, date of access: August 2023.}. As we see from the table, already for batches of size 1,000, $\partial$SGP4 offers a significant speed-up, when run in batch mode, on both CPU and GPU, compared to the standard SGP4 program. In particular, as we see, the GPU implementation can reach a factor of ~85$\times$ speed-up compared to the standard SGP4. This level of speed-up factor holds promise for applications such as all-versus-all conjunction screening, where computational speed-up can yield substantial benefits.

\begin{table}[ht]
\centering
\caption{Execution Times for SGP4, $\partial$SGP4 on CPU, and \texorpdfstring{$\partial$}{partial}SGP4 on GPU}
\begin{tabular}{lcccccccccc}
\toprule
\multirow{3}{*}{\textbf{Model}} & \multicolumn{10}{c}{\textbf{Run Time (ms) per Event}} \\
& \multicolumn{2}{c}{\textbf{Single Event}} & \multicolumn{2}{c}{\textbf{10,000}} & \multicolumn{2}{c}{\textbf{100,000}} & \multicolumn{2}{c}{\textbf{1,000,000}} & \multicolumn{2}{c}{\textbf{10,000,000}} \\
\midrule
\multirow{1}{*}{SGP4} & 0.002 & & 21.11 & & 211.1 & & 2,111 & & 21,110 \\
\midrule
\multirow{1}{*}{$\partial$SGP4 (1 CPU)} & 0.67 & & 4.52 & & 44.6 & & 460 & & 7,920 & \\
\midrule
\multirow{1}{*}{$\partial$SGP4 (GPU)} & 4.22 & & 7.8 & & 8 & & 29 & & 249 & \\
\bottomrule
\end{tabular}
\label{tab:dSGP4_speed}
\end{table}
\section{Machine Learning \texorpdfstring{$\partial$}{partial}SGP4}
\label{sec:mldSGP4}
\subsection{Methods}
\label{sec:mldSGP4_methods}
As detailed in Section~\ref{sec:automatic_differentiation} the propagator can be seamlessly integrated into complex machine learning pipelines, offering a differentiable program for end-to-end training using techniques like stochastic gradient descent. While general perturbation propagators like SGP4 boast fast propagation times, their low accuracy limits their applicability, especially in scenarios like collision avoidance.

\begin{figure}[htb!]
    \centering
    \begin{tabular}{c|c}
    \begin{subfigure}{0.37\textwidth}
        \includegraphics[width=\textwidth]{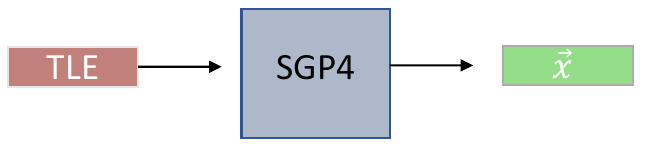}
        \caption{SGP4 model.}
        \label{fig:subfig_sgp4_model}
    \end{subfigure}
    &
    \begin{subfigure}{0.37\textwidth}
        \includegraphics[width=\textwidth]{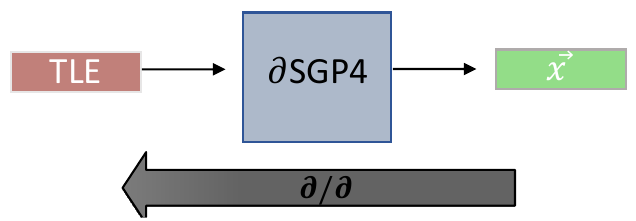}
        \caption{$\partial$SGP4 model.}
        \label{fig:subfig_dSGP4_model}
    \end{subfigure}
    \\
    \hline
    \begin{subfigure}{0.47\textwidth}
        \includegraphics[width=\textwidth]{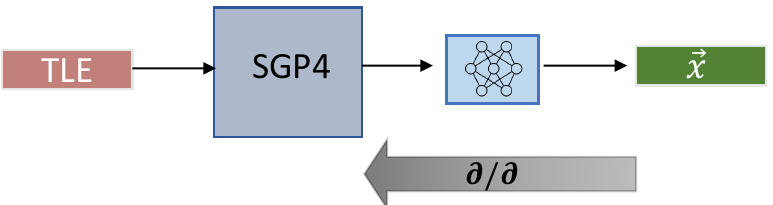}
        \caption{ML-SGP4 model.}
        \label{fig:subfig_sgp4_ml_model}
    \end{subfigure}
    &
    \begin{subfigure}{0.47\textwidth}
        \includegraphics[width=\textwidth]{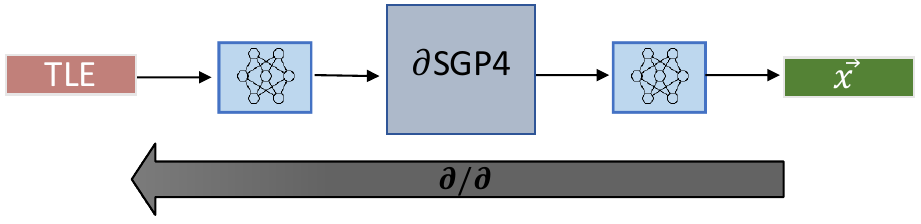}
        \caption{ML-$\partial$SGP4 model.}
        \label{fig:subfig_dSGP4_ml_model_}
    \end{subfigure}
    \end{tabular}
    \caption{Schematic illustration of SGP4, ML-SGP4, $\partial$SGP4, and ML-$\partial$SGP4 models.}
    \label{fig:dSGP4_and_ML-dSGP4_models}
\end{figure}

This section introduces ML-$\partial$SGP4, a paradigm where a differentiable propagator ($\partial$SGP4) is integrated with two neural networks to correct its inputs and outputs. In Figure~\ref{fig:subfig_sgp4_model} and ~\ref{fig:subfig_dSGP4_model}, respectively, both standard SGP4 and $\partial$SGP4 propagations are illustrated. While in Figure~\ref{fig:subfig_dSGP4_model} and ~\ref{fig:subfig_dSGP4_ml_model_}, respectively, the combination between ML and both SGP4 and $\partial$SGP4 are shown.

The ML-$\partial$SGP4 model represents a new hybrid differentiable propagator that retains the computational advantages of SGP4 while enabling training and improvement with accurate simulated or observed data. The model can be configured to return either SGP4 predictions or activate input and output corrections using trained neural networks.

\begin{figure}[htb!]
     \centering
    \includegraphics[width=\textwidth]{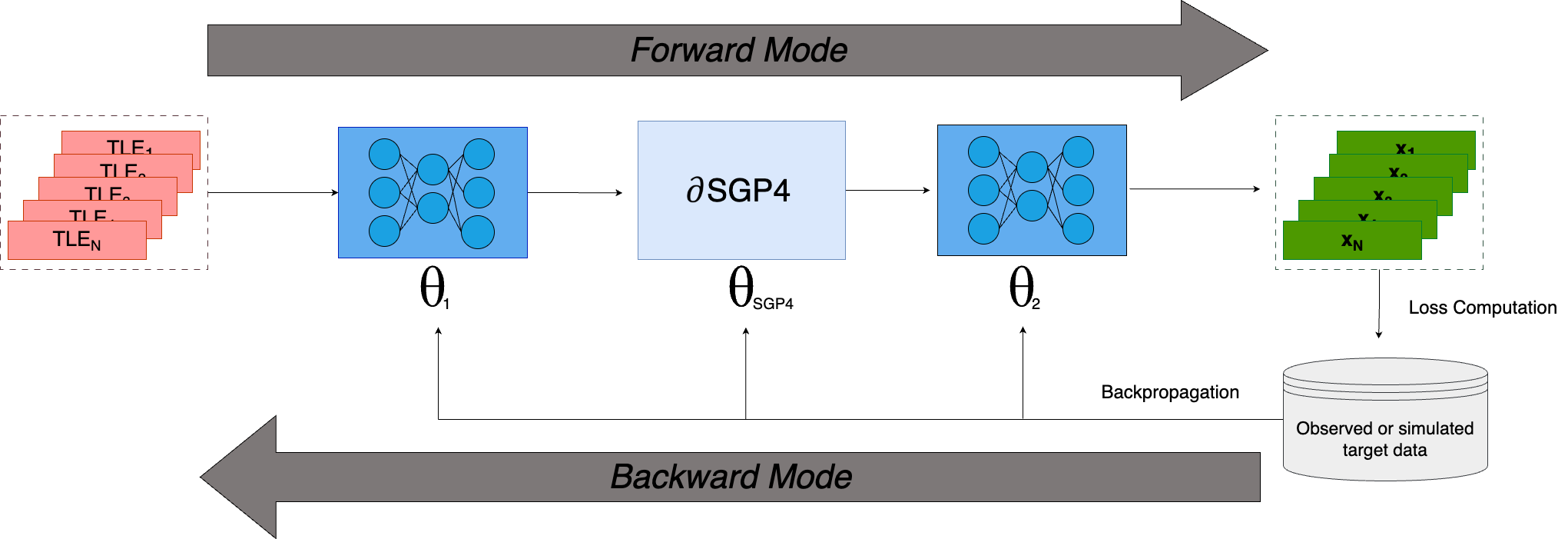}
\caption{Schematic illustration of ML-$\partial$SGP4 model, in both \textit{forward} and \textit{backward} mode.}
\label{fig:dsgp4_forward_and_backward}
\end{figure}

To better discuss the ML-$\partial$SGP4 setup shown in Figure~\ref{fig:dSGP4_and_ML-dSGP4_models}, let us call $\vec{x}_0$ the input of ML-$\partial$SGP4 (i.e., the TLE elements), $t$ the propagation time, and $\vec{x}$ its output, and let us assume that the objective is to minimize the error between the ML-$\partial$SGP4 propagation and the propagation of a high precision propagator, $\vec{x}_{\textrm{HPOP}}$ (i.e., a numerical integrator that accounts for all forces involved in the satellite motion). In this case, the objective is to correct ML-$\partial$SGP4 to better match the accurate propagator, hence, the function to minimize will be a function of the state, time, and parameters (both of the networks and also of $\partial$SGP4). Without loss of generality, let's consider as an objective function the squared norm between the target state and the $\partial$SGP4 output, then the objective is to find the parameters (i.e., weights and biases) of the two networks, as well as the parameters of SGP4 that minimize the loss:
\begin{equation}
\mathcal{P}:
    \begin{cases}
        \mbox{given:} & \vec{x}_0, t \\
        \mbox{find:} & \vec{\theta}_1, \vec{\theta}_{\textrm{SGP4}}, \vec{\theta}_2 \\
        \mbox{s.t. min:} & J(\vec{x}(\vec{x}_0, t, \vec{\theta}_1, \vec{\theta}_{\textrm{SGP4}}, \vec{\theta}_2)) = ||\vec{x}-\vec{x}_{\textrm{HPOP}}||^2
    \end{cases}
    \text{,}
\end{equation}

where $\vec{\theta}_1$ indicates the parameters of the input neural network, $\vec{\theta}_2$ of the output one, and  $\vec{\theta}_{\textrm{SGP4}}$ the SGP4 parameters. 
Owing to the differentiability of the neural networks and $\partial$SGP4, we can use gradient descent to find the local minimum of the loss function by iteratively updating the parameters of the neural networks and $\partial$SGP4: 
\begin{align}
    \begin{split}
        &\vec{\theta}_{1,i+1}=\vec{\theta}_{1,i}-\eta_1 \nabla_{\vec{\theta}_1}J\\        &\vec{\theta}_{\textrm{SGP4},i+1}=\vec{\theta}_{\textrm{SGP4},i}-\eta_{\textrm{SGP4}} \nabla_{\vec{\theta}_{\textrm{SGP4}}}J\\
        &\vec{\theta}_{2,i+1}=\vec{\theta}_{2,i}-\eta_2 \nabla_{\vec{\theta}_2}J\\
    \end{split}
    \text{.}
\end{align}
This setup can be easily extended for multiple propagation points, for instance, by taking the mean squared error as a loss function, and by setting up a stochastic gradient descent algorithm to update the parameters. A schematic illustration of this architecture is shown in Figure~\ref{fig:dsgp4_forward_and_backward}. As it can be seen, the model is first run in \textit{forward} mode: in this way, from a set of TLEs, a set of propagated states can be found. Then, in \textit{backward} mode, the loss between these states and simulated (e.g. HPOP simulations) or observed states (e.g. ephemeris data) is computed via automatic differentiation, and the partials of the loss w.r.t. both neural network parameters and $\partial$SGP4 can be used to update the models, until convergence. 
The advantage of ML-$\partial$SGP4 lies in its capacity to learn both the neural network weights and SGP4 parameters, therefore offering a flexible solution. This differs from ML-SGP4, where only the output of the SGP4 propagator is corrected, as explored in prior research~\cite{san2018hybrid}.  In the next section, we will delve into some numerical experiments to showcase the use of ML-$\partial$SGP4 and to highlight the comparison between ML-$\partial$SGP4 and the ML-SGP4.

\subsection{Experiments}
\label{sec:mldSGP4_experiments}

This section presents experiments conducted to assess the accuracy of the ML-$\partial$SGP4 propagator and compare it with the uncorrected SGP4 baseline and a neural network corrector that only adjusts SGP4 outputs~\cite{san2018hybrid}. 

The dataset comprises SpaceX Starlink Two-Line Elements (TLEs) and ephemeris data available on Space-Track\footnote{\url{https://www.space-track.org/}, date of access: April 2023.}. The ephemeris data includes a set of 1,519 Starlink satellites, in the period from the 18\textsuperscript{th} of February 2023, at 11:38:42 UTC, to the 21\textsuperscript{st} of February 2023, at 11:38:42 UTC, with a time resolution of 60 seconds. This data is provided by SpaceX using their numerical orbital propagator. 
We also collect the TLE data for the same objects during the period of the 17\textsuperscript{th} and 18\textsuperscript{th} of February, before the date of the first available ephemeris. We then use $\partial$SGP4 to propagate those TLEs at the same times as the available ephemeris data, and we treat each TLE as a separate observation. 
This generates a dataset of as many entries as the number of TLEs for each object, times the number of time stamps in the ephemeris data, multiplied by the total number of satellites: that is, a dataset with about 6.5 million rows and 6 columns (three for the position vector and three for the velocity vector). 

We split the training, validation, and test sets at the TLE level, making sure that the data propagated from a certain TLE belongs to the same set and is not shared among multiple sets: this is to avoid the model essentially learning to interpolate the data. We use 69\% of the dataset as training, 16\% as validation, and 15\% as testing. 
Moreover, we make sure that the test set covers ranges of altitude that are unseen in the training and validation set: this resulted in 225 Starlink satellites below 539 km (while all the other satellites above that altitude have been used in training and validation). 
We train the model with the architecture shown on the right side of Figure~\ref{fig:dSGP4_and_ML-dSGP4_models}, by using two neural networks (each with three hidden layers of 35 neurons each), using 30 epochs, a learning rate of 3$\times10^{-3}$, mean squared error (MSE) as a loss, and Adam as optimizer. The total number of learnable parameters was 3,454. 

To have a fair comparison, we compare this model with a neural network output corrector that uses a similar number of learnable parameters (i.e., 3596): resulting from a feed-forward neural network with four hidden layers of size 32, plus the six correction factors for the output corrections. In both this and the ML-$\partial$SGP4 case, the networks only learn corrections to the given inputs and/or outputs.

\begin{figure}
  \begin{minipage}{0.333\textwidth}
    \centering
    \includegraphics[width=\linewidth]{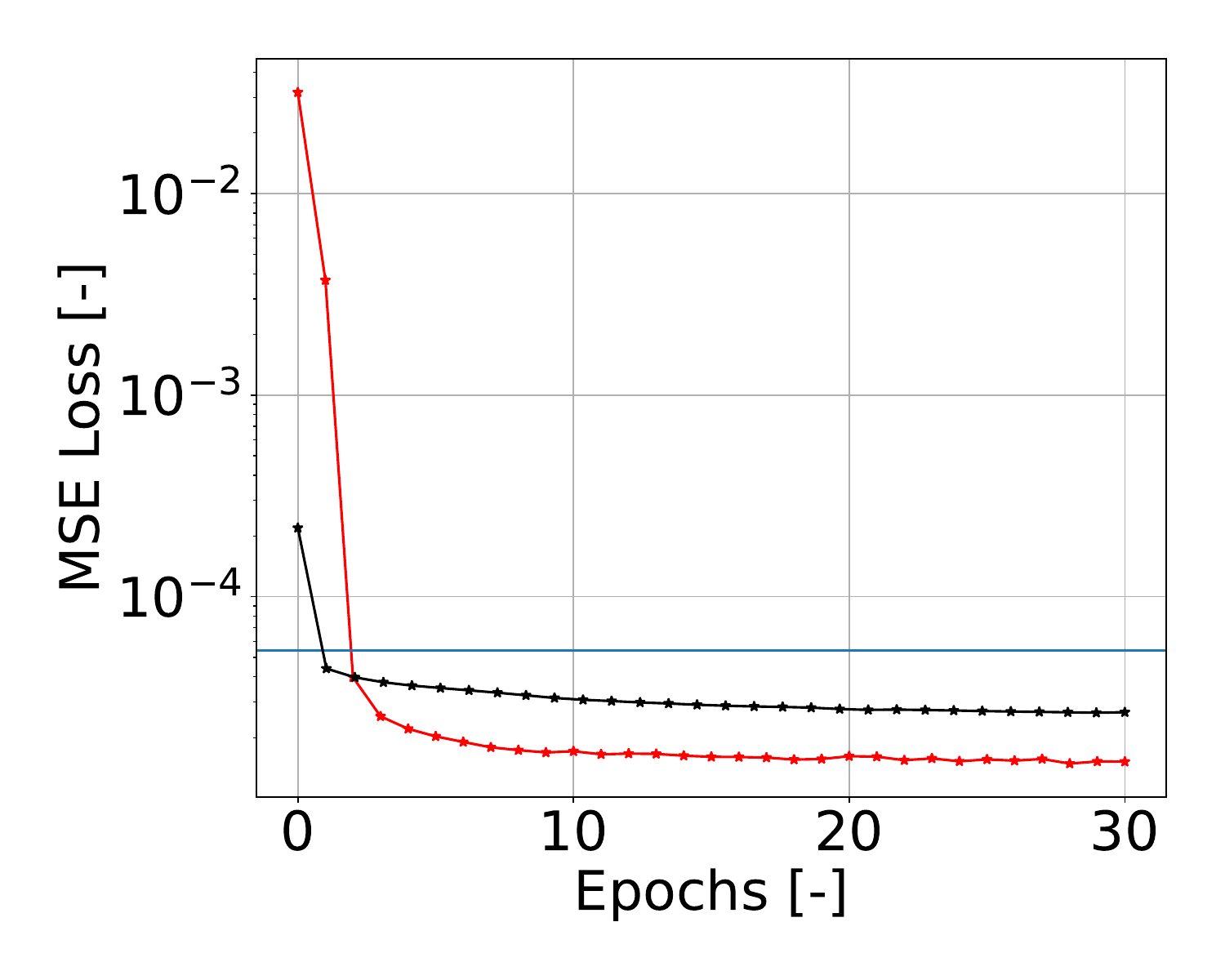}
  \end{minipage}%
  \begin{minipage}{0.333\textwidth}
    \centering
    \includegraphics[width=\linewidth]{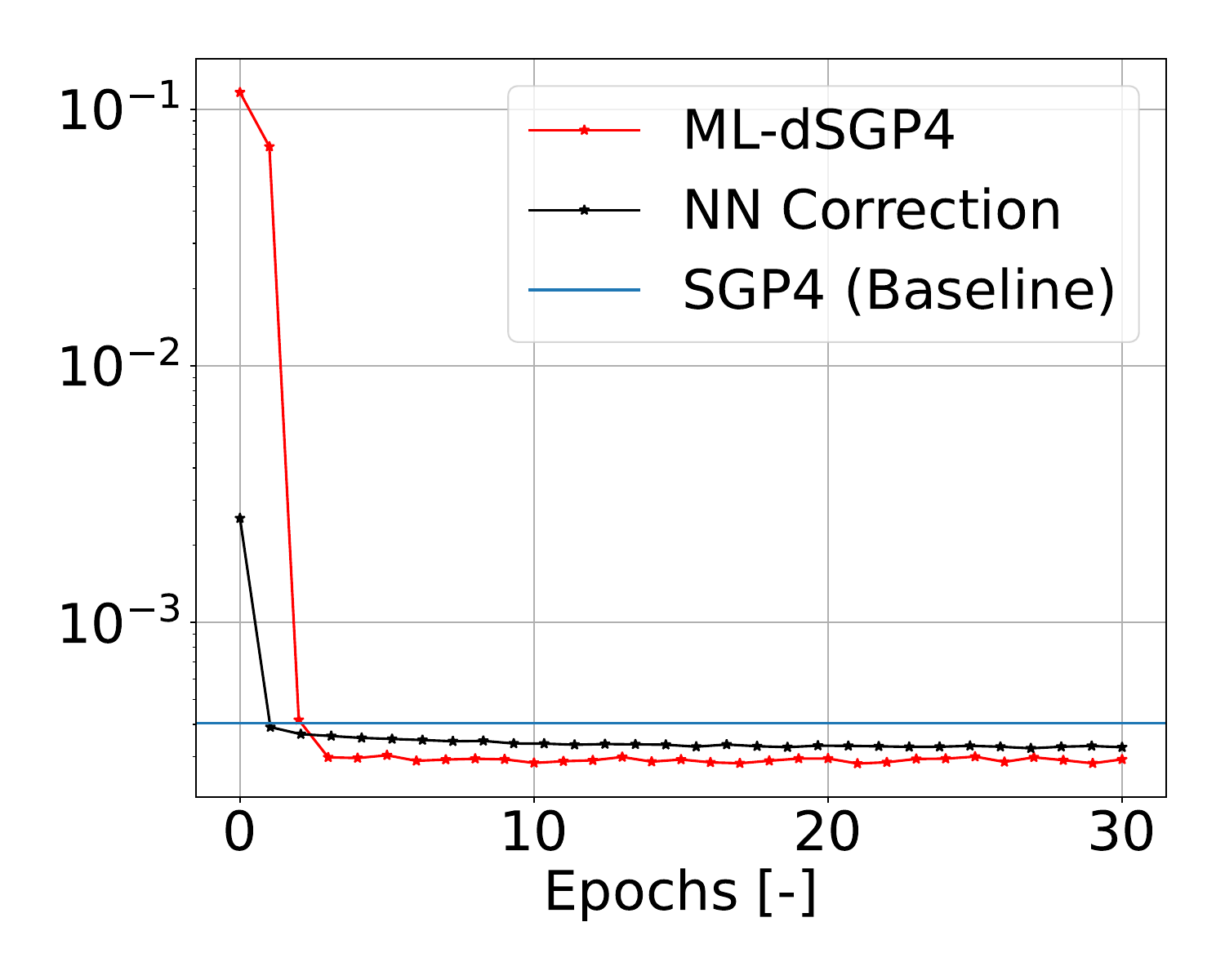}
  \end{minipage}%
  \begin{minipage}{0.333\textwidth}
    \centering
    \includegraphics[width=\linewidth]{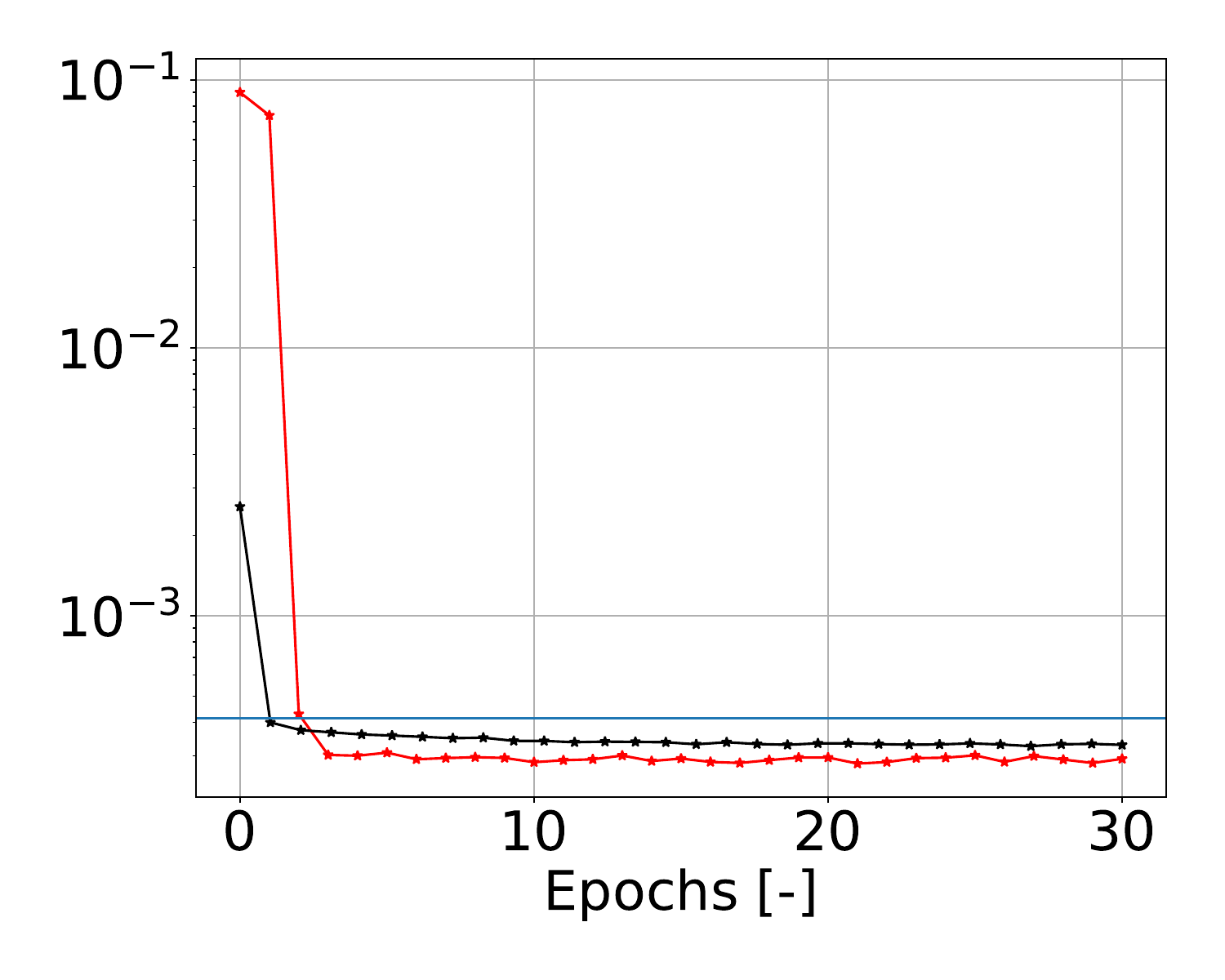}
  \end{minipage}
  \caption{Training, validation, and test losses mean squared error as a function of epochs.}
  \label{fig:train_valid_test_losses}
  \end{figure}

In Figure~\ref{fig:train_valid_test_losses}, we show the loss of the training, validation, and test data as a function of epochs for both the ML-$\partial$SGP4 and the neural network corrector model. As we observe, in all three datasets, both models manage to outperform the SGP4 baseline model, and ML-$\partial$SGP4 manages to achieve a higher accuracy than the neural network corrector, thereby confirming the added advantages of the differentiable setup that can also learn input corrections. 

In Table~\ref{tab:best_l}, we show the final losses of both models, compared to the baseline results: to produce the table, the best models at validation were chosen for the neural network corrector and ML-$\partial$SGP4. 
As already mentioned, the test set was in a set of altitudes (and therefore also dynamics) completely unseen during training or validation. Nonetheless, the ML-$\partial$SGP4 model manages to achieve accuracies in the test set that are in the same range of values as the validation ones: this confirms that the model successfully starts to learn a differential correction between the low-accuracy SGP4 and the high-precision orbital propagation model. This paves the way for a mixed strategy where low-precision propagation with $\partial$SGP4 can be combined with observed or simulated data to learn correction models that then enable to perform fast and more accurate predictions of orbital states. Besides, this is also combined with the possibility of performing these computations in batches and leveraging parallelization via CPU or GPU. Finally, in this experiment we did not learn any internal $\partial$SGP4 model correction factors: however, the same framework can also be leveraged to allow $\partial$SGP4 internal parameters to be learned, thereby providing further prediction accuracy enhancements.
\begin{table}[!ht]
\centering
\begin{tabular}{|c|c|c|c|c|c|}
\hline
& & State [-] & Pos [km] & Vel [km/s] \\
\hline
& Training & 5.41179$\times10^{-5}$ & 46.6 &  0.0508\\
SGP4 (baseline)~\cite{hoots1980spacetrack}& Validation & 0.000404 & 100.32 & 0.1097 \\
& Test & 0.0004153 &  107.42 & 0.1176\\
\hline
& Training & 2.69409$\times10^{-5}$ & 33.36 & 0.0363 \\
ML-SGP4~\cite{san2018hybrid} & Validation & 0.000323 & 81.27 & 0.0888\\
& Test & 0.000326 & 88.66 & 0.0969 \\
\hline
& Training & 1.75807$\times10^{-5}$ & 25.01 & 0.0276 \\
ML-$\partial$SGP4 & Validation & 0.000262 & 64.02 & 0.0704\\
& Test & 0.000256 & 71.25 & 0.0783 \\
\hline
\end{tabular}
\caption{Baseline, ML-SGP4, and ML-$\partial$SGP4 MSE values for the normalized state, and the position and velocity root MSE after 25 epochs, for training, validation, and test datasets.}
\label{tab:best_l}
\end{table}

The objective of this experiment is to show that even with only three days of ephemeris data, the ML-$\partial$SGP4 model can be leveraged to increase the accuracy of the standard SGP4 propagator while retaining its computational advantages. Feeding more ephemeris data, the model can further enhance its accuracy, potentially starting to be competitive with high-precision numerical propagators in terms of precision, while retaining the advantages of general perturbation techniques in terms of speed. This might also be very appealing for operators of constellations that possess a large amount of high-precision data.

\section{Satellite Orbit Determination}
\label{sec:satellite_orbit_determination}
\subsection{Methods}
\label{sec:sod_methods}
Satellite orbit determination consists of the accurate determination of spacecraft states starting from a dynamical model of the forces acting on the satellite (e.g. SGP4) and available observations (e.g. obtained through ground-based instruments). The objective is to combine a simulated environment and raw data to provide a good estimate of the satellites' position and velocity. The orbit determination processes both in general and in relation to SGP4 are detailed in Vallado's publications~\cite{vallado2001fundamentals, vallado2008sgp4}. 

In the aforementioned works, either finite differences or analytical schemes are proposed as a means to find the Jacobian matrix needed for solving the least squares problem. However, in our case, we will show that this is not necessary, since the differentiability of $\partial$SGP4 enables the computation of those quantities via AD, which turns out to be very advantageous, as it avoids extra computation and/or additional errors. 

The detailed problem setup is that we have $N$ observations, $\{\hat{x}_i\}_{i=1}^N$, at different observation times, $\{t_i\}_{i=1}^N$. Our objective is to find the state at a certain target time, $t_T$, that best fits the set of observations. Effectively, this means finding the state at time $t_T$: $\vec{x}_T$, which minimizes the (weighted) squared error between the observed and modeled states at the observation times. This problem can be described as follows:
\begin{equation}
\mathcal{P}:
    \begin{cases}
        \mbox{given:} & \{\hat{x}_i, t_i\}_{i=1}^N, \ t_T \\
        \mbox{find:} & \vec{x}_T \\
        \mbox{s.t. min:} & \sum_{i=1}^N |(\vec{x}_i(\vec{x}_T, t_T)-\hat{x}_i)w_i|^2
    \end{cases}
    \text{,}
\end{equation}
where $w_i$ are the observation weights. The modeled states $\vec{x}_i$ at the observed times $t_i$, can be found via the nonlinear propagation of $\vec{x}_T$ at $t_T$, until $t_i$, using $\partial$SGP4:
\begin{equation}
    \vec{x}_i = \textrm{$\partial$SGP4}(\vec{x}_T, t_i-t_T)
\end{equation}
The above problem is equivalent to a weighted least-squares problem, whose minimum can be sought starting from an initial guess and setting up an update rule that depends on the Jacobian of the modeled states at observation times, with respect to the state at the target time~\cite{montenbruck2002satellite}. 

Hence, starting from an initial guess $\vec{x}_T^0$, we can setup an iteration scheme:
\begin{equation}
    \vec{x}_T^{k+1}=\vec{x}_T^{k}+\delta \vec{x}
    \text{,}
\end{equation}
where: 
\begin{equation}
    \delta \vec{x}=-(A^TWA)^{-1}A^TW\vec{b}\text{,}
\end{equation}
with $W$ being the weight matrix, $\vec{b}_i=(\vec{x}_{i}^{k}-\hat{x}_{i})$ and $A_i$ is the Jacobian of the $i$-th modeled state with respect to the target state:
\begin{equation}
    A_i=\dfrac{\partial \vec{x}_i}{\partial \vec{x}_T}\bigg|_{\vec{x}_T^k}
    \text{.}
    \label{eq:a_matrix_od}
\end{equation}
Owing to the differentiability of $\partial$SGP4, the partial derivative terms in Eq.~\eqref{eq:a_matrix_od} can now be exactly computed via automatic differentiation.
In case more than one observations are available, then the $A$, $W$ matrices and $\vec{b}$ vector can be found by stacking each observation: $A^TWA=[A_0^TWA_0,\dots, A_N^TWA_N]^T$, $A^TW\vec{b}=[A_0^TW\vec{b},\dots, A_N^TW\vec{b}_N]$. In case no weights are assigned for the observations, then the weight matrix corresponds to the identity matrix.

\subsection{Experiments}
\label{sec:sod_experiments}
Using the procedure highlighted in Section~\ref{sec:satellite_orbit_determination}, we setup an experiment to show the usage of $\partial$SGP4 in an orbit determination problem. In our analysis, we assume to have a set of TLEs as observations, and we use SGP4 as a nonlinear orbital propagation model: our goal is to set up the problem as a nonlinear least squares problem, in which we want to find a TLE for a given target epoch that best fits the observations, using the SGP4 model.  This experiment is also documented in one of the tutorials of the open-source package. As we already mentioned, three essential components of satellite orbit determination are a set of observations, a target time at which we desire to extract the satellites' state, and an initial guess to setup the iterative procedure. We download the set of TLEs corresponding to Sentinel-2A satellite in the period from 1st of June to 8th of June 2022, from the Space-Track\footnote{\url{https://www.space-track.org/}, date of access: April 2023.} public catalog. This corresponds to a set of 26 TLEs, with a maximum time difference of about 10 hours, and a minimum time difference of about 3 hours: we use these TLEs as observations. Then, we choose the target epoch, at which we would like to extract the best fit for the TLE, one day after the final observation, which corresponds to the 8th of June 2022, at 21:25:48.337248 UTC. The initial guess for the TLE elements is found from the average state of the 26 TLEs propagated at the target epoch via $\partial$SGP4. This corresponds to the following initial state: $\vec{r}=[4.5995\times10^{2}, 2.4107\times10^{3}, 6.7259\times10^{3}]^T$ km, $\vec{v}=[4.3698, 5.5925, -2.2983]^T$ km/s, which results in the nominal target TLE (i.e., $\vec{x}_T^0$) shown in Tab.~\ref{tab:nominal_tle_initial}. By propagating the nominal TLE at the observation times and comparing the norm of the residuals w.r.t. the actual observation, we get a total residual norm in the state of 2915.6622, with residual norms ranging from a maximum of 900 km in position and 1 km/s in velocity, to a minimum of 100 km in position and 0.1 km/s in velocity for the closest in time. We use this nominal TLE as an initial guess for the differential corrector. After convergence, the TLE in Tab.~\ref{tab:nominal_tle_converged} is found, with a total residual norm in the state (norm of the sum of all residuals) of 1.2112, which corresponds to residuals that range from a minimum of 12.6 m in position and 0.5 m/s in velocity, and to a maximum of 513.6 m in position and 1.4 m/s in velocity. The difference between this example and previous satellite orbit determination applications shown in literature is that in this case, we used $\partial$SGP4 automatic differentiation to precisely compute the partial derivatives needed to build the $A$ matrix of Eq.~\eqref{eq:a_matrix_od}, therefore avoiding the local truncation errors due to finite differencing~\cite{vallado2008sgp4}.
\begin{table}[!ht]
\centering
\begin{tabular}{c}
    \hline
  1 \ 40697U \ 15028A \ \ \ 22159.89292057 \ \ .00000111 \ \ 00000-0 \ \ 59112-4 \ 0  \ \ 9998 \\
  \hline
  2 \ 40697 \ \ 98.5685 \  234.7917 \  0000491 \  348.2227 \ 119.9277 \ 14.31085333362518 \\
  \hline
    \end{tabular}
    \caption{initial nominal TLE.}
    \label{tab:nominal_tle_initial}
\end{table}

\begin{table}[!ht]
\centering
\begin{tabular}{c}
    \hline
  1 \ 40697U \ 15028A \ \ \ 22159.89292057 \ \ .00000111 \ \ 00000-0 \ \ -19814-4 \ 0  \ \ 9998 \\
  \hline
  2 \ 40697 \ \ 98.5716 \  234.7932 \  0000011 \  348.1175 \ 120.0022 \ 14.30817171362512 \\
  \hline
    \end{tabular}
    \caption{TLE after convergence.}
    \label{tab:nominal_tle_converged}
\end{table}

\section{Conclusions}
\label{sec:conclusions}
In this work, we introduce $\partial$SGP4, a differentiable version of the SGP4 model implemented in PyTorch, representing a significant advancement in space situational awareness. Our research work yields several considerations. 

First of all, the differentiability features of the model enable the accurate computation of partial derivatives through automatic differentiation. This finds applications in gradient-based optimization, orbit determination, state transition matrix computation, sensitivity matrix computation, covariance propagation, and covariance transformation.

Moreover, we discussed the seamless integration of $\partial$SGP4 with state-of-the-art machine learning backpropagation algorithms. This opens new avenues for orbital propagation models. We, therefore, introduce ML-$\partial$SGP4, a machine learning SGP4 model to enhance SGP4 propagations to better match accurate simulated or observed data. During training, both $\partial$SGP4 parameters and neural network parameters can be jointly optimized backpropagating through both the $\partial$SGP4 model and the neural networks, enhancing prediction accuracy to align more closely with the data. In this way, one can maintain the same SGP4 model efficiency and speed, while improving its accuracy and enabling parallelization over CPU and GPU. We show how the new propagation model is able to outperform both the SGP4 baseline and black-box neural network correctors on a sample of around 1,500 Starlink satellites.

Furthermore, the PyTorch implementation of $\partial$SGP4 facilitates straightforward parallelization on both CPU and GPUs, thereby allowing the processing of batches of TLE data at the same time. This acceleration holds promise for applications that display bottlenecks in terms of computational costs, such as many-to-many conjunction screening in an increasingly congested low-Earth orbit.

Throughout our paper, we present several experiments that delve into each of these applications, providing quantitative results that underscore the practical significance and versatility of $\partial$SGP4 in advancing space situational awareness and orbital prediction. Furthermore, we also release the $\partial$SGP4 code and associated tests and tutorials as an open-source Github repository under the ESA organization.

\section{Acknowledgments}
\label{sec:acknowledgments}
We gratefully acknowledge Dr. T.S. Kelso for sharing his expertise and providing valuable suggestions, particularly regarding the validation of our work. 
\bibliography{references}

\end{document}